\documentclass[conference]{IEEEtran}
\usepackage[utf8]{inputenc}
\usepackage{amsmath,amssymb,amsfonts}
\usepackage{graphicx}
\usepackage{textcomp}
\usepackage{xcolor}
\usepackage{tabto}
\usepackage{dirtytalk}
\usepackage{graphicx}
\usepackage{setspace}
\usepackage{mathtools}
\usepackage{accents}
\usepackage{enumitem}
\usepackage{hyperref}
\usepackage{longtable}
\usepackage{fixmath}
\usepackage{wrapfig}
\usepackage{color}
\usepackage{algorithm,algpseudocode}
\usepackage{comment}
\usepackage{subfig}
\usepackage{soul}
\usepackage{tabularx}
\usepackage{multirow}
\usepackage{lipsum}
\usepackage{graphicx}
\usepackage{siunitx}
\usepackage{amssymb}
\usepackage{array}

\usepackage{times}
\usepackage{xcolor}
\usepackage{soul}
\usepackage{amsmath}
\usepackage{mathtools}
\usepackage{accents}
\usepackage{amssymb}
\usepackage{enumitem}
\usepackage{hyperref}
\usepackage{enumitem}
\usepackage{longtable}
\usepackage{fixmath}
\usepackage{amsmath}
\usepackage{wrapfig}

\def\BibTeX{{\rm B\kern-.05em{\sc i\kern-.025em b}\kern-.08em
    T\kern-.1667em\lower.7ex\hbox{E}\kern-.125emX}}

\begin{document}

\title{Improved Classification Based on Deep Belief Networks}

%\author{}

\author{\IEEEauthorblockN{Jaehoon Koo}
\IEEEauthorblockA{\textit{Northwestern University} \\
Evanston, IL, USA \\
jaehoonkoo2018@u.northwestern.edu} \vspace{.5cm}

\and 

\IEEEauthorblockN{Diego Klabjan}
\IEEEauthorblockA{\textit{Northwestern University} \\
Evanston, IL, USA \\
d-klabjan@northwestern.edu} \vspace{.5cm}
}

\maketitle

\begin{abstract}
For better classification generative models are used to initialize the model and model features before training a classifier. Typically it is needed to solve separate unsupervised and supervised learning problems. Generative restricted Boltzmann machines and deep belief networks are widely used for unsupervised learning. We developed several supervised models based on DBN in order to improve this two-phase strategy. Modifying the loss function to account for expectation with respect to the underlying generative model, introducing weight bounds, and multi-level programming are applied in model development. The proposed models capture both unsupervised and supervised objectives effectively. The computational study verifies that our models perform better than the two-phase training approach.
\end{abstract}

\begin{IEEEkeywords}
deep learning, neural networks, classification
\end{IEEEkeywords}

%\section{To do}
%1. Add a t-test results.

%The difference in the classification error between our best model (BL for NI, FFN-DBNOPT for ISOLET) and 2-phase are statistically significant since their p-values are 0.03, 0.01, 0.03 for NI 20\%, 30\%, and 40\% datasets, and 0.07 for ISOLET. %For MNIST, our best model, FFN-DBNOPT, obtains a lower mean classification error than 2-phase with the same standard deviation. %We agree that the average improvement is not drastic however in deep learning it is very hard to make substantial improvements due to overwhelming research in recent years. We stress again that while our improvements are not substantial, they are consistent and not just flux or noise.
%For NI, we use the same amount of data in training as that used in \cite{Salama2011}.%, which is why we assume that 40\% of the full data is large enough to train our models.

%2. Possible outlets
%\begin{itemize}
%  \item IEEE ICDM 2019, due: Full paper submissions: June 5, 2019  
%  \item IEEE BigData 2019, due: Electronic submission of full papers: August 19, 2019
%  \item SIAM International Conference on Data Mining (SDM19), due: October 12, 2018
%\end{itemize}

\section{Introduction}

Restricted Boltzmann machine (RBM), an energy-based model to define an input distribution, is widely used to extract latent features before classification. Such an approach combines unsupervised learning for feature modeling and supervised learning for classification. Two training steps are needed. The first step, called pre-training, is to model features used for classification. This can be done by training RBM that captures the distribution of input. The second step, called fine-tuning, is to train a separate classifier based on the features from the first step~\cite{Larochelle2012}. This two-phase training approach for classification is also used for deep networks. Deep belief networks (DBN) are built with stacked RBMs, and trained in a layer-wise manner~\cite{Hinton2006}. Two-phase training based on a deep network consists of DBN and a classifier on top of it.

The two-phase training strategy has three possible problems. 1) It requires two training processes; one for training RBMs and one for training a classifier. 2) It is not guaranteed that the modeled features in the first step are useful in the classification phase since they are obtained independently of the classification task. 3) It is an effort to decide which classifier is the best for each problem. Therefore, there is a need for a method that can conduct feature modeling and classification concurrently~\cite{Larochelle2012}.

To resolve these problems, recent papers suggest to transform RBM to a model that can deal with both unsupervised and supervised learning. Since RBM calculate the joint and conditional probabilities, the suggested prior models combine a generative and discriminative RBM. Consequently, this hybrid discriminative RBM is trained concurrently for both objectives by summing the two contributions~\cite{Larochelle2012,Larochelle2008}. In a similar way a self-contained RBM for classification is developed by applying the free-energy function based approximation to RBM, which was used for a supervised learning method, reinforcement learning~\cite{Elfwing2015}. However, these approaches are limited to transforming RBM that is a shallow network. 

In this study, we developed alternative models to solve a classification problem based on DBN. Viewing the two-phase training as two separate optimization problems, we applied optimization modeling techniques in developing our models. Our first approach is to design new objective functions. We design an expected loss function based on $p(h|x)$ built by DBN and the loss function of the classifier. Second, we introduce constraints that bound the DBN weights in the feed-forward phase. The constraints keep a good representation of input as well as regularize the weights during updates. Third, we applied bilevel programming to the two-phase training method. The bilevel model has a loss function of the classifier in its objective function but it constrains the DBN values to the optimal to phase-1. This model searches possible optimal solutions for the classification objective only where DBN objective solutions are optimal. 

Our main contributions are several classification models combining DBN and a loss function in a coherent way. In the computational study we verify that the suggested models perform better than the two-phase method. 

\section{Literature Review}
 
The two-phase training strategy has been applied to many classification tasks on different types of data. Two-phase training with RBM and support vector machine (SVM) has been explored in classification tasks on images, documents, and network intrusion data ~\cite{Xing2005,Norouzi2009,Salama2011,Dahl2012}. Logistic regression replacing SVM has been explored~\cite{Mccallum2006,Cho2011}. Gehler et al.~\cite{Gehler2006} used the 1-nearest neighborhood classifier with RBM to solve a document classification task. Hinton and Salakhutdinov~\cite{Hinton2006} suggested DBN consisting of stacked RBMs that is trained in a layer-wise manner. Two-phase method using DBN and deep neural network has been studied to solve various classification problems such as image and text recognition  ~\cite{Hinton2006,Bengio2007,Sarikaya2014}. Recently, this approach has been applied to motor imagery classification in the area of brain–computer interface~\cite{Lu2017}, biomedical research, classification of Cytochrome P450 1A2 inhibitors and non-inhibitors~\cite{Yu2017}, and web spam classification that detects web pages deliberately created to manipulate search rankings ~\cite{Li2018}. All these papers rely on two distinct phases, while our models assume a holistic view of both aspects.

Many studies have been conducted to improve the problems of two-phase training. Most of the research has been focused on transforming RBM so that the modified model can achieve generative and discriminative objectives at the same time. Schmah et al.~\cite{Schmah2009} proposed a discriminative RBM method, and subsequently classification is done in the manner of a Bayes classifier. However, this method cannot capture the relationship between the classes since the RBM of each class is trained separately. Larochelle et al.~\cite{Larochelle2012,Larochelle2008} proposed a self-contained discriminative RBM framework where the objective function consists of the generative learning objective $p(x,y)$, and the discriminative learning objective, $p(y|x)$. Both distributions are derived from RBM. Similarly, a self-contained discriminative RBM method for classification is proposed~\cite{Elfwing2015}. The free-energy function based approximation is applied in the development of this method, which is initially suggested for reinforcement learning. This prior paper relying on RBM conditional probability while we handle general loss functions. Our models also hinge on completely different principles. 

\section{Background}
\label{gen_inst}
\paragraph{Restricted Boltzmann Machines} 
RBM is an energy-based probabilistic model, which is a restricted version of Boltzmann machines (BM) that is a log-linear Markov Random Field. It has visible nodes $x$ corresponding to input and hidden nodes $h$ matching the latent features. The joint distribution of the visible nodes $x \in \mathbb{R}^J $ and hidden variable $h \in \mathbb{R}^I$ is defined as
		\begin{equation*}
		\begin{aligned}
		& p(x,h) = \frac{1}{Z} e^{-E(x,h)}, \: E(x,h) = - h W x - c h - b x
		\end{aligned}
		\end{equation*}
where $W \in \mathbb{R}^{I \times J}, \: b \in \mathbb{R}^{J}$, and  $c \in \mathbb{R}^{I}$ are the model parameters, and  \(Z\) is the partition function. Since units in a layer are independent in RBM, we have the following form of conditional distributions:
		\begin{equation*}
		\begin{aligned}
		& p(h|x) = \prod_{i=1}^{I} p(h_i | x), \: p(x|h) = \prod_{j=1}^{J} p(x_j | h). \\
		\end{aligned}
		\end{equation*}  
For binary units where $x \in \{0,1\}^{J}$ and $h \in \{0,1\}^I$, we can write $ p(h_i=1|h)= \sigma(c_i+W_ix)$ and $p(x_j=1|h)=\sigma(b_j + W_j x)$ where $\sigma()$ is the sigmoid function. In this manner RBM with binary units is an unsupervised neural network with a sigmoid activation function. The model calibration of RBM can be done by minimizing negative log-likelihood through gradient descent. RBM takes advantage of having the above conditional probabilities which enable to obtain model samples easier through a Gibbs sampling method. Contrastive divergence (CD) makes Gibbs sampling even simpler: 1) start a Markov chain with training samples, and 2) stop to obtain samples after k steps. It is shown that CD with a few steps performs effectively~\cite{Bengio2009,Hinton2002}.

\paragraph{Deep Belief Networks}
DBN is a generative graphical model consisting of stacked RBMs. Based on its deep structure DBN can capture a hierarchical representation of input data. Hinton et al. (2006) introduced DBN with a training algorithm that greedily trains one layer at a time. Given visible unit $x$ and $\ell$ hidden layers the joint distribution is defined as~\cite{Bengio2009,Hinton2006a}
		\begin{equation*}
		\begin{aligned}
		& p(x, h^1, \cdots, h^\ell) = p(h^{\ell-1}, h^\ell) \left( \prod_{k=1}^{\ell-2} p(h^k|h^{k+1}) \right) p(x|h^1)   .
		\end{aligned}
		\end{equation*}
Since each layer of DBN is constructed as RBM, training each layer of DBN is the same as training a RBM. 
 
Classification is conducted by initializing a network through DBN training~ \cite{Bengio2007,Hinton2006a}. A two-phase training can be done sequentially by: 1) pre-training, unsupervised learning of stacked RBM in a layer-wise manner, and 2) fine-tuning, supervised learning with a classifier. Each phase requires solving an optimization problem. Given training dataset $D = \{(x^{(1)},y^{(1)}),\dots, (x^{(|D|)},y^{(|D|)})\}$ with input $x$ and label $y$, the pre-training phase solves the following optimization problem at each layer $k$
		\begin{equation*}
		\begin{aligned}
		& \underset{ \substack{ \theta_k}}{\text{min}} \hspace{0.5cm} \frac{1}{|D|} \sum_{i =1}^{|D|} \left[ - log \: p(x_k^{(i)};\theta_k) \right] \\
		\end{aligned}
		\end{equation*}
where $\theta_k = (W_k,b_k,c_k)$ is the RBM model parameter that denotes weights, visible bias, and hidden bias in the energy function, and $x_k^{(i)}$ is visible input to layer $k$ corresponding to input $x^{(i)}$. Note that in layer-wise updating manner we need to solve $\ell$ of the problems from the bottom to the top hidden layer. For the fine-tuning phase we solve the following optimization problem
		\begin{equation} \label{eq:fine}
		\begin{aligned}
		& \underset{ \substack{\phi}}{\text{min}} \hspace{0.5cm} \frac{1}{|D|} \sum_{i =1}^{|D|} \left[\mathcal{L}(\phi; y^{(i)}, h(x^{(i)})) \right] \\ 
		\end{aligned}
		\end{equation}
where $\mathcal{L} ()$ is a loss function, $h$ denotes the final hidden features at layer $\ell$, and $\phi$ denotes the parameters of the classifier. Here for simplicity we write $h(x^{(i)}) = h(x_{\ell}^{(i)})$. When combining DBN and a feed-forward neural networks (FFN) with sigmoid activation, all the weights and hidden bias parameters among input and hidden layers are shared for both training phases. Therefore, in this case we initialize FFN by training DBN. 

\section{Proposed Models}

We model an expected loss function for classification. Considering classification of two phase method is conducted on hidden space, the probability distribution of the hidden variables obtained by DBN is used in the proposed models. The two-phase method provides information about modeling parameters after each phase is trained. Constraints based on the information are suggested to prevent the model parameters from deviating far from good representation of input. Optimal solution set for unsupervised objective of the two-phase method is good candidate solutions for the second phase. Bilevel model has the set to find optimal solutions for the phase-2 objective so that it conducts the two-phase training at one-shot.  

\paragraph{DBN Fitting Plus Loss Model}
We start with a naive model of summing pre-trainning and fine-tuning objectives. This model conducts the two-phase training strategy simultaneously; however, we need to add one more hyperparameter $\rho$ to balance the impact of both objectives. The model (DBN+loss) is defined as
        \begin{equation*}
		\begin{aligned}
		& \underset{ \substack{\theta_{L}, \theta_{DBN}}}{\text{min}} \hspace{0.1cm} \mathbb{E}_{\textbf{y,x}} [\mathcal{L}(\theta_{L}; \textbf{y}, h(\textbf{x}))] + \rho \:\mathbb{E}_{\textbf{x}}[ -\: log \: p(\textbf{x};\theta_{DBN})     ] 
		\end{aligned}
		\end{equation*}
and empirically based on training samples $D$,
		\begin{equation} \label{eq:RBM+loss}
		\begin{aligned}
		& \underset{ \substack{\theta_{L}, \theta_{DBN}}}{\text{min}} \hspace{0.0cm} \frac{1}{|D|} \sum_{i =1}^{|D|} \left[\mathcal{L}(\theta_{L}; y^{(i)}, h(x^{(i)})) - \rho \: log \: p(x^{(i)};\theta_{DBN}) \right]
		\end{aligned}
		\end{equation}
where $\theta_{L}, \theta_{DBN}$ are the underlying parameters. Note that $\theta_{L} = \phi$ from (\ref{eq:fine}) and $\theta_{DBN} = (\theta_{k})_{k=1}$. This model has already been proposed 
%by Larochelle et al. (2008, 2012) 
if the classification loss function is based on the RBM conditional distribution~\cite{Larochelle2012,Larochelle2008}. 

%\paragraph{Expected loss model}
%We propose an expected loss model based on conditional distribution \(p(h|x)\) obtained by DBN. This model conducts classification on the hidden space. Since it minimizes the expected loss, it should be more robust and thus it should yield better accuracy on data not observed. The mathematical model that minimizes the expected loss function is defined as  
%		\begin{equation*}
%		\begin{aligned}
%		& \underset{ \substack{\theta_{Loss}, \theta_{DBN}}}{\text{min}} \hspace{0.5cm} \mathbb{E}_{\textbf{y,h}|\textbf{x}} [\mathcal{L}(\theta_{Loss}; \textbf{y}, h (\theta_{DBN} ;\textbf{x}))] 
%		\end{aligned}
%		\end{equation*}
%and empirically based on training samples \(D\),
%		\begin{equation*}
%		\begin{aligned}
%		& \underset{ \substack{\theta_{Loss}, \theta_{DBN}}}{\text{min}} \hspace{0.5cm} \frac{1}{|D|} \sum_{i =1}^{|D|} \left[  \sum_{h} p(h|x^{(i)}) \mathcal{L}(\theta_{Loss}; y^{(i)}, h(\theta_{DBN};x^{(i)})) \right]. \\
%		\end{aligned}
%		\end{equation*}       
%With notation \(h(\theta_{DBN};x^{(i)}) = h(x^{(i)})\) we explicitly show the dependency of \(h\) on \(\theta_{DBN}\).
\paragraph{Expected Loss Model with DBN Boxing}
We first design an expected loss model based on conditional distribution $p(h|x)$ obtained by DBN. This model conducts classification on the hidden space. Since it minimizes the expected loss, it should be more robust and thus it should yield better accuracy on data not observed. The mathematical model that minimizes the expected loss function is defined as  
		\begin{equation*}
		\begin{aligned}
		& \underset{ \substack{\theta_{L}, \theta_{DBN}}}{\text{min}} \hspace{0.5cm} \mathbb{E}_{\textbf{y,h}|\textbf{x}} [\mathcal{L}(\theta_{L}; \textbf{y}, h (\theta_{DBN} ;\textbf{x}))] 
		\end{aligned}
		\end{equation*}
and empirically based on training samples $D$,
		\begin{equation*}
		\begin{aligned}
		& \underset{ \substack{\theta_{L}, \theta_{DBN}}}{\text{min}} \hspace{0.1cm} \frac{1}{|D|} \sum_{i =1}^{|D|} \left[  \sum_{h} p(h|x^{(i)}) \mathcal{L}(\theta_{L}; y^{(i)}, h(\theta_{DBN};x^{(i)})) \right]. \\
		\end{aligned}
		\end{equation*}       
With notation $h(\theta_{DBN};x^{(i)}) = h(x^{(i)})$ we explicitly show the dependency of $h$ on \(\theta_{DBN}\). We modify the expected loss model by introducing a constraint that sets bounds on DBN related parameters with respect to their optimal values. This model has two benefits. First, the model keeps a good representation of input by constraining parameters fitted in the unsupervised manner. Also, the constraint regularizes the model parameters by preventing them from blowing up while being updated. Given training samples $D$ the mathematical form of the model (EL-DBN) reads
		\begin{equation*}
		\begin{aligned}
		& \underset{ \substack{\theta_{L}, \theta_{DBN}}}{\text{min}} \hspace{0.0cm} \frac{1}{|D|} \sum_{i =1}^{|D|} \left[  \sum_{h} p(h|x^{(i)}) \mathcal{L}(\theta_{L}; y^{(i)}, h(\theta_{DBN};x^{(i)})) \right] \\
		& \hspace{0.37cm} \text{s.t.} \hspace{0.48cm} |\theta_{DBN}-\theta_{DBN}^{*}| \leq \delta \\
		\end{aligned}
		\end{equation*}
where  $\theta_{DBN}^{*}$ are the optimal DBN parameters and $\delta$ is a hyperparameter. This model needs a pre-training phase to obtain the DBN fitted parameters.
\paragraph{Expected Loss Model with DBN Classification Boxing}
Similar to the DBN boxing model, this expected loss model has a constraint that the DBN parameters are bounded by their optimal values at the end of both phases. This model regularizes parameters with those that are fitted in both the unsupervised and supervised manner. Therefore, it can achieve better accuracy even though we need an additional training to the two-phase trainings. Given training samples $D$ the model (EL-DBNOPT) reads
		\begin{equation} \label{eq:EL-DBNOPT}
		\begin{aligned}
		& \underset{ \substack{\theta_{L}, \theta_{DBN}}}{\text{min}} \hspace{0.1cm} \frac{1}{|D|} \sum_{i =1}^{|D|} \left[  \sum_{h} p(h|x^{(i)}) \mathcal{L}(\theta_{L}; y^{(i)}, h(\theta_{DBN};x^{(i)})) \right] \\
		& \hspace{0.37cm} \text{s.t.} \hspace{0.48cm} |\theta_{DBN}-\theta_{DBN-OPT}^{*}| \leq \delta \\
		\end{aligned}
		\end{equation}
where  $\theta_{DBN-OPT}^{*}$ are the optimal values of DBN parameters after two-phase training and $\delta$ is a hyperparameter.

\paragraph{Feed-forward Network with DBN Boxing}
We also propose a model based on boxing constraints where FFN is constrained by DBN output. The mathematical model (FFN-DBN) based on training samples $D$ is
		\begin{equation} \label{eq:FFN-DBN}
		\begin{aligned}
		& \underset{ \substack{\theta_{L}, \theta_{DBN}}}{\text{min}} \hspace{0.5cm} \frac{1}{|D|} \sum_{i =1}^{|D|} \left[ \mathcal{L}(\theta_{L}; y^{(i)}, h(\theta_{DBN};x^{(i)})) \right] \\
		& \hspace{0.5cm} \text{s.t.} \hspace{0.8cm} |\theta_{DBN}-\theta_{DBN}^{*}| \leq \delta. \\
		\end{aligned}
		\end{equation}
		
\paragraph{Feed-forward Network with DBN Classification Boxing}
Given training samples $D$ this model (FFN-DBNOPT), which is a mixture of (\ref{eq:EL-DBNOPT}) and (\ref{eq:FFN-DBN}), reads
		\begin{equation*}
		\begin{aligned}
		& \underset{ \substack{\theta_{L}, \theta_{DBN}}}{\text{min}} \hspace{0.5cm} \frac{1}{|D|} \sum_{i =1}^{|D|} \left[ \mathcal{L}(\theta_{L}; y^{(i)}, h(\theta_{DBN};x^{(i)})) \right] \\
		& \hspace{0.5cm} \text{s.t.} \hspace{0.8cm}  |\theta_{DBN}-\theta_{DBN-OPT}^{*}| \leq \delta. \\
		\end{aligned}
		\end{equation*}    
 
\paragraph{Bilevel Model} We also apply bilevel programming to the two-phase training method. This model searches optimal solutions to minimize the loss function of the classifier only where DBN objective solutions are optimal. Possible candidates for optimal solutions of the first level objective function are optimal solutions of the second level objective function. This model (BL) reads
		\begin{equation*}
		\begin{aligned}
		& \underset{\substack{\theta_{L},\theta_{DBN}^{*}}} {\text{min}}  \hspace{0.5cm} \mathbb{E}_{\textbf{y,x}} [\mathcal{L}(\theta_{L}; \textbf{y},h(\theta_{DBN}^{*} ;\textbf{x}))]\\
		& \hspace{0.5cm} \text{s.t.} \hspace{0.8cm}  \theta_{DBN}^{*} = \operatorname*{arg\,min}_{\theta_{DBN}} \hspace{0.2cm} \mathbb{E}_{\textbf{x}} [-log \: p(\textbf{x};\theta_{DBN})]
		\end{aligned}
		\end{equation*}
and empirically based on training samples,
		\begin{equation*}
		\begin{aligned}
		& \underset{\substack{\theta_{L},\theta_{DBN}^{*}}} {\text{min}}  \hspace{0.2cm} \frac{1}{|D|} \sum_{i =1}^{|D|} \left[ \mathcal{L}(\theta_{L}; y^{(i)}, h(\theta_{DBN}^{*};x^{(i)})) \right]\\
		& \hspace{0.35cm} \text{s.t.} \hspace{0.5cm} \theta_{DBN}^{*} = \operatorname*{arg\,min}_{\theta_{DBN}} \hspace{0.2cm} \frac{1}{|D|} \sum_{i =1}^{|D|} \left[ - log \: p(x^{(i)};\theta_{DBN}) \right].
		\end{aligned}
		\end{equation*}
One of the solution approaches to bilevel programming is to apply Karush–Kuhn–Tucker (KKT) conditions to the lower level problem. After applying KKT to the lower level, we obtain
		\begin{equation*}
		\begin{aligned}
		& \underset{\substack{\theta_{L},\theta_{DBN}^{*}}}{\text{min}} \hspace{0.5cm} \mathbb{E}_{\textbf{y,x}} [\mathcal{L}(\theta_{L}; \textbf{y},h(\theta_{DBN}^{*}  ;\textbf{x}))] \\
		& \hspace{0.4cm} \text{s.t.} \hspace{0.8cm}  \nabla_{\theta_{DBN}} \mathbb{E}_{\textbf{x}} [-log \: p(\textbf{x};\theta_{DBN})|_{{\theta_{DBN}^{*}}} ]= 0. \\ 
		\end{aligned}
		\end{equation*} 
Furthermore, we transform this constrained problem to an unconstrained problem with a quadratic penalty function:
		\begin{equation} \label{eq:BL}
		\begin{aligned}
		& \underset{\substack{\theta_{L},\theta_{DBN}^{*}}}{\text{min}} \hspace{0.3cm}  \mathbb{E}_{\textbf{y,x}} [\mathcal{L}(\theta_{L}; \textbf{y},h(\theta_{DBN}^{*} ;\textbf{x}))] + \\
		& \hspace{1.5cm} \frac{\mu}{2} || \nabla_{\theta_{DBN}} \mathbb{E}_{\textbf{x}} [-log \: p(\textbf{x};\theta_{DBN})] |_{\theta_{DBN}^{*}} ||^2 
		\end{aligned}
		\end{equation}
where $\mu$ is a hyperparameter. The gradient of the objective function is derived in the appendix.

\section{Computational Study}

To evaluate the proposed models classification tasks on three datasets were conducted: the MNIST hand-written images \footnote{\href{http://yann.lecun.com/exdb/mnist/}{http://yann.lecun.com/exdb/mnist/}}, the KDD'99 network intrusion dataset (NI)\footnote{\href{http://kdd.ics.uci.edu/databases/kddcup99/kddcup99.html}{http://kdd.ics.uci.edu/databases/kddcup99/kddcup99.html}}, and the isolated letter speech recognition  dataset (ISOLET) \footnote{\href{https://archive.ics.uci.edu/ml/datasets/ISOLET}{https://archive.ics.uci.edu/ml/datasets/ISOLET}}. The experimental results of the proposed models on these datasets were compared to those of the two-phase method. 

In FFN, the sigmoid functions in the hidden layers and the softmax function in the output layer were chosen with negative log-likelihood as a loss function of the classifiers. We selected the hyperparameters based on the settings used in \cite{Wang2017}, which were fine-tuned. We first implemented the two-phase method with DBNs of 1, 2, 3 and 4 hidden layers to find the best configuration for each dataset, and then applied the best configuration to the proposed models.

Implementations were done in Theano using GeForce GTX
TITAN X. The mini-batch gradient descent algorithm was used to solve the optimization problems of each model. To calculate the gradients of each objective function of the models Theano's built-in functions, 'theano.tensor.grad', was used. We denote by DBN-FFN the two-phase approach. 

\subsection{MNIST}

The task on the MNIST is to classify ten digits from 0 to 9 given by $28\times28$ pixel hand-written images. The dataset is divided in 60,000 samples for training and validation, and 10,000 samples for testing. The hyperparameters are set as: there are 1,000 hidden units in each layer; the number of pre-training epochs per layer is 100 with the learning rate of 0.01; the number of fine-tuning epochs is 300 with the learning rate of 0.1; the batch size is 10; and $\rho$ in the DBN+loss and $\mu$ in the BL model are diminishing during epochs. Note that DBN+Loss and BL do not require pre-training.

DBN-FFN with three-hidden layers of size, 784-1000-1000-1000-10, was the best, and subsequently we compared it to the proposed models with the same size of the network. We computed the means of the classification errors and their standard deviations for each model averaged over 5 random runs. In each table, we stressed in bold the best three models with ties broken by deviation. In Table 1, the best mean test error rate was achieved by FFN-DBNOPT, 1.32\%. Furthermore, the models with the DBN classification constraints, EL-DBNOPT and FFN-DBNOPT, perform similar to, or better than the two-phase method. This shows that DBN classification boxing constraints regularize the model parameters by keeping a good representation of input.

\begin{table}[t]
%\label{MNIST-table}
\centering
\begin{tabular}{lcc}
    \hline
             & Mean   & Sd. \\
    \hline
    DBN-FFN  & \textbf{1.33\%} & \textbf{0.03\%} \\
    DBN+loss & 1.84\% & 0.14\% \\
    EL-DBN   & 1.46\% & 0.05\% \\
    EL-DBNOPT& \textbf{1.33\%} & \textbf{0.04\%} \\
    FFN-DBN  & 1.34\% & 0.04\% \\
   FFN-DBNOPT&\textbf{1.32\%}& \textbf{0.03\%} \\
    BL       & 1.85\% & 0.07\%\\
    \hline
\end{tabular}
\caption{Classification errors with respect to the best DBN structure for the MNIST.}
\end{table}

\subsection{Network Intrusion}

The classification task on NI is to distinguish between normal and bad connections given the related network connection information. The preprocessed dataset consists of 41 input features and 5 classes, and 4,898,431 examples for training and 311,029 examples for testing. The experiments were conducted on 20\%, 30\%, and 40\% subsets of the whole training set, which were obtained by stratified random sampling. The hyperparameters are set as: there are 15 hidden units in each layer; the number of pre-training epochs per layer is 100 with the learning rate of 0.01; the number of fine-tuning epochs is 500 with the learning rate of 0.1; the batch size is 1,000; and $\rho$ in the DBN+loss and $\mu$ in the BL model are diminishing during epochs.

On NI the best structure of DBN-FFN was 41-15-15-5 for all three datasets. Table 2 shows the experimental results of the proposed models with the same network as the best DBN-FFN. BL performed the best in all datasets, achieving the lowest mean classification error without the pre-training step. The difference in the classification error between our best model, BL, and DBN-FFN is statistically significant since the p-values are 0.03, 0.01, and 0.03 for 20\%, 30\%, and 40\% datasets, respectively. This showed that the model being trained concurrently for unsupervised and supervised purpose can achieve better accuracy than the two-phase method. Furthermore, both EL-DBNOPT and FFN-DBNOPT yield similar to, or lower mean error rates than DBN-FFN in all of the three subsets. 

\begin{table}[t]
%  \label{NI-tab}
  \centering
  \begin{tabular}{lcccc}
    \hline
             & \multicolumn{2}{c}{20\% dataset} & \multicolumn{2}{c}{30\% dataset}  \\
    \cline{2-5}     
             &Mean & Sd. & Mean & Sd.  \\ 
    \hline
    DBN-FFN  & 8.14\% & 0.12\% & 8.18\% & 0.12\% \\
    DBN+loss & \textbf{8.07\%} & \textbf{0.06\%} & \textbf{8.13\%} & \textbf{0.09\%} \\
    EL-DBN   & 8.30\% & 0.09\% & 8.27\% & 0.07\% \\
    EL-DBNOPT& 8.14\% & 0.14\% & 8.15\% & 0.15\% \\
    FFN-DBN  & 8.17\% & 0.09\% & 8.20\% & 0.08\% \\
   FFN-DBNOPT& \textbf{8.07\% }& \textbf{0.12\%}& \textbf{8.12\%} & \textbf{0.11\%} \\
    BL       & \textbf{7.93\%} & \textbf{0.09\%} & \textbf{7.90\%} & \textbf{0.11\%} \\
    \hline
  \end{tabular}
  \begin{tabular}{lcc}
    \hline
             & \multicolumn{2}{c}{40\% dataset} \\
    \cline{2-3}
             & Mean & Sd. \\ 
    \cline{1-3}
    DBN-FFN  & 8.06\% & 0.02\% \\
    DBN+loss & \textbf{8.05\%} & \textbf{0.05\%} \\
    EL-DBN   & 8.29\% & 0.14\% \\
    EL-DBNOPT& 8.08\% & 0.10\% \\
    FFN-DBN  & 8.07\% & 0.11\% \\
   FFN-DBNOPT& \textbf{7.95\%} & \textbf{0.11\%} \\
    BL       & \textbf{7.89\%} & \textbf{0.10\%} \\
    \cline{1-3}
  \end{tabular}  
\caption{Classification errors with respect to the best DBN structure for NI}
\end{table}

\subsection{ISOLET}
The classification on ISOLET is to predict which letter-name was spoken among the 26 English alphabets given 617 input features of the related signal processing information. The dataset consists of 5,600 for training, 638 for validation, and 1,559 examples for testing. The hyperparameters are set as: there are 1,000 hidden units in each layer; the number of pre-training epochs per layer is 100 with the learning rate of 0.005; the number of fine-tuning epochs is 300 with the learning rate of 0.1; the batch size is 20; and $\rho$ in the DBN+loss and $\mu$ in the BL model are diminishing during epochs.

In this experiment the shallow network performed better than the deep network; 617-1000-26 was the best structure for DBN-FFN. One possible reason for this is its small size of training samples. EL models performed great for this instance. EL-DBNOPT achieved the best mean classification error, tied with FFN-DBNOPT. With the same training effort, EL-DBN achieved a lower mean classification error and smaller standard deviation than the two-phase method, DBN-FFN. Considering a relatively small sample size of ISOLET, EL shows that it yields better accuracy on unseen data as it minimizes the expected loss, i.e., it generalizes better. In this data set, p-value is 0.07 for the difference in the classification error between our best model, FFN-DBNOPT, and DBN-FFN.

\begin{table}[t]
%\label{MNIST-table}
\centering
\begin{tabular}{lcc}
    \hline
             & Mean   & Sd. \\
    \hline
    DBN-FFN  & 3.94\% & 0.22\% \\
    DBN+loss & 4.38\% & 0.20\% \\
    EL-DBN   & \textbf{3.91\%} & \textbf{0.18\%} \\
    EL-DBNOPT& \textbf{3.75\%} & \textbf{0.14\%} \\
    FFN-DBN  & 3.94\% & 0.19\% \\
   FFN-DBNOPT&\textbf{3.75\%}& \textbf{0.13\%} \\
    BL       & 4.43\% & 0.18\%\\
    \hline
\end{tabular}
\caption{Classification errors with respect to the best DBN structure for ISOLET.}
\end{table}

\section{Conclusions}

DBN+loss performs better than two-phase training DBN-FFN only in one instance. Aggregating two unsupervised and supervised objectives without a specific treatment is not effective. Second, the models with DBN boxing, EL-DBN and FFN-DBN, do not perform better than DBN-FFN in almost all datasets. Regularizing the model parameters with unsupervised learning is not so effective in solving a supervised learning problem. Third, the models with DBN classification boxing, EL-DBNOPT and FFN-DBNOPT, perform better than DBN-FFN in almost all of the experiments. FFN-DBNOPT is consistently one of the best three performers in all instances. This shows that classification accuracy can be improved by regularizing the model parameters with the values trained for unsupervised and supervised purpose. One drawback of this approach is that one more training phase to the two-phase approach is necessary. Last, BL shows that one-step training can achieve a better performance than two-phase training. Even though it worked in one instance, improvements to current BL can be made such as applying different solution search algorithms, supervised learning regularization techniques, or different initialization strategies.

\bibliographystyle{ieeetr}
\bibliography{main}

\appendix

\section{Approximation of DBN Probability in the Proposed Models}
DBN defines the joint distribution of the visible unit $x$ and the $\ell$ hidden layers, $h^1, h^2, \cdots, h^{\ell}$ as 
\begin{equation*}
\begin{aligned}
& p(x, h^1, \cdots, h^\ell) = p(h^{\ell-1}, h^\ell) \left( \prod_{k=0}^{\ell-2} p(h^k|h^{k+1}) \right) 
\end{aligned}
\end{equation*}
with $h^0 = x$.
\paragraph{DBN Fitting Plus Loss Model}
From Eq. (\ref{eq:RBM+loss}), $p(x)$ in the second term of the objective function is approximated as 		
\begin{equation*}
\begin{aligned}
& p(x;\theta_{DBN}) = \sum_{h^1, h^2, \cdots, h^{\ell}} p(x, h^1, \cdots, h^\ell) \approx \sum_{h^1} p (x, h^1).
\end{aligned}
\end{equation*}
\paragraph{Expected Loss Models}
$p(h|x)$ in the objective function is approximated as 		
\begin{equation*}
\begin{aligned}
%& p(h^{\ell} | h^{\ell-1}) =  \sum_{h^1, h^2, \cdots, h^{\ell-2}} p(h^{\ell} | h^{\ell-1}, h^{\ell-2}, \cdots, x) = \sum_{h^1, h^2, \cdots, h^{\ell-2}}. \\
& p(h^{\ell} | x) \approx p( h^{\ell} | x, h^1, \cdots, h^{\ell}) \\
& \hspace{1.1cm}    =  \frac {p(h^{\ell}, h^{\ell-1}, \cdots, h^{1}, x)}{p(h^{\ell-1}, h^{\ell-2}, \cdots, h^{1}, x)} \\
& \hspace{1.1cm}    = \frac{p(h^{\ell-1}, h^\ell) \left( \prod_{k=0}^{\ell-2} p(h^k|h^{k+1}) \right)}{ p(h^{\ell-2}, h^{\ell-1}) \left( \prod_{k=0}^{\ell-3} p(h^k|h^{k+1}) \right)} \\
& \hspace{1.1cm}    = \frac{p(h^{\ell-1}, h^\ell) p(h^{\ell-2}|h^{\ell-1}) }{ p(h^{\ell-2}, h^{\ell-1})} \\
& \hspace{1.1cm}    = \frac{p(h^{\ell-1}, h^\ell) p(h^{\ell-2},h^{\ell-1}) }{ p(h^{\ell-2}, h^{\ell-1})p(h^{\ell-1})} \\
& \hspace{1.1cm}    = p(h^{\ell} | h^{\ell-1}).\\ 
\end{aligned}
\end{equation*}
\paragraph{Bilevel Model}
From Eq. (\ref{eq:BL}), $\nabla_{\theta_{DBN}} log \: p(x)$ in the objective function is approximated for $i = 0, 1, \cdots, \ell$ as
\begin{equation} \label{eq:BLapp}
\begin{aligned}
& \left[\nabla_{\theta_{DBN}} log \: p(x) \right]_{i} =\frac{ \partial \: log \: p(x)}{\partial \: \theta_{DBN}^{i}} \\
& \hspace{2.2cm}= \frac{ \partial \: log \: \left(\sum_{h^1, h^2, \cdots, h^{\ell}} p(x, h^1, h^2, \cdots, h^{\ell}) \right)}{\partial \: \theta_{DBN}^{i}} \\ 
& \hspace{2.2cm} \approx \frac{ \partial \: log \: (\sum_{h^{i+1}} p(h^i, h^{i+1}))}{\partial \: \theta_{DBN}^{i}}  \\
\end{aligned}
\end{equation}
where \(\theta_{DBN} = (\theta_{DBN}^0, \theta_{DBN}^2, \cdots, \theta_{DBN}^{i}, \cdots \theta_{DBN}^{\ell}) \). The gradient of this approximated quantity is then the Hessian matrix of the underlying RBM. 

\section{Derivation of the Gradient of the Bilevel Model}
We write the approximated $|| \nabla_{\theta_{DBN}} - log \: p(x) ||^2$ at the layer $i$ as
\begin{equation*}
\begin{aligned}
& || [\nabla_{\theta_{DBN}} - log \: p(x)]_i ||^2 \approx || \frac{ \partial \: - log \: (\sum_{h^{i+1}} p(h^i, h^{i+1}))}{\partial \: \theta_{DBN}^{i}} ||^2 \\
& \hspace{0.0cm} =  \Bigg[  \left(\frac{\partial -log \: p(h^i) }{\partial \theta_{11}^i} \right)^2 + \left(\frac{\partial -log \: p(h^i) }{\partial \theta_{12}^i} \right)^2 + \\
& \hspace{0.6cm} \cdots + \left(\frac{\partial -log \: p(h^i) }{\partial \theta_{nm}^i} \right)^2 \Bigg]
\end{aligned}
\end{equation*}
where $m$ and $n$ denote dimensions of $h^i$ and $h^{i+1}$ and $\theta_{pq}^i$ denotes the $p^{th}$ and $q^{th}$ component of the $\theta_{DBN}^i$. The gradient of the approximated $|| \nabla_{\theta_{DBN}} - log \: p(x) ||^2$ at the layer $i$ is 
\begin{equation*}
\begin{aligned}
& \hspace{0cm} \frac{\partial \:  }{\theta_{pq}^{i}} \left(  \sum_{p,q} \left(\frac{\partial -log \: p(h^i) }{\partial \theta_{pq}^i} \right)^2  \right)  \\
& = 2 \Bigg[  \left(\frac{\partial -log \: p(h^i) }{\partial \theta_{11}^i} \right) \left(\frac{\partial^2 -log \: p(h^i) }{\partial \theta_{11}^i \theta_{pq}^i} \right) + \\
& \hspace{0.9cm} \left(\frac{\partial -log \: p(h^i) }{\partial \theta_{12}^i} \right) \left(\frac{\partial^2 -log \: p(h^i) }{\partial \theta_{12}^i \partial \theta_{pq}^i} \right) + \\
& \hspace{0.5cm} \cdots  + \left(\frac{\partial -log \: p(h^i) }{\partial \theta_{pq}^i} \right) \left(\frac{\partial^2 -log \: p(h^i) }{\partial \theta_{pq}^i \partial \theta_{pq}^i} \right) +  \\
& \hspace{0.5cm} \cdots + \left(\frac{\partial -log \: p(h^i) }{\partial \theta_{nm}^i} \right) \left(\frac{\partial^2 -log \: p(h^i) }{\partial \theta_{nm}^i \theta_{pq}^i} \right)   \Bigg] \quad \\ 
\end{aligned}
\end{equation*}
for  $p = 1, ... n, q = 1, ...m$. This shows that the gradient of the approximated $ || \nabla_{\theta_{DBN}} - log \: p(x) ||^2 $ in (\ref{eq:BL}) is then the Hessian matrix times the gradient of the underlying RBM. The stochastic gradient of \( -log \: p(x)\) of RBM with binary input \(x\) and hidden unit \(h\) with respect to \(\theta_{DBN} w_{pq}\) is 
\begin{equation*}
\begin{aligned}
& \frac{\partial RBM}{\partial w_{pq}} = p(h_p = 1 | x) x_q - \sum_{x} p(x) p(h_p = 1 |x) x_q
\end{aligned}
\end{equation*}
where \(RBM\) denotes \(-log \: p(x)\) \cite{Fischer2012}. We derive the Hessian matrix with respect to \(w_{pq}\) as
\begin{equation*}
\begin{aligned}
& \frac{\partial^2 RBM}{\partial w_{pq}^2} \\
& = \frac{\partial}{w_{pq}} [p(h_p = 1|x) x_q)] - \sum_x \frac{\partial}{w_{pq}} [p(x) p(h_p = 1|x) x_q)] \\
& \hspace{0.0cm} = \sigma (\widetilde{net_p}) (1- \sigma(\widetilde{net_p})) x_q^2 - \sum_x [\frac{\partial p(x) }{\partial w_{pq}} p(h_p = 1|x) x_q + \\
& \hspace{0.5cm} p(x)\sigma (\widetilde{net_p}) (1- \sigma(\widetilde{net_p})) x_q^2], \\ 
& \frac{\partial^2 RBM}{\partial w_{pk} \partial w_{pq}} \\
& = \frac{\partial}{w_{pk}} [p(h_p = 1|x) x_q)] -  \frac{\partial}{w_{pk}} [ \sum_x p(x) p(h_p = 1|x) x_q)] \\
& \hspace{0cm} = \sigma (\widetilde{net_p}) (1- \sigma(\widetilde{net_p})) x_q x_k - \sum_x [\frac{\partial p(x) }{\partial w_{pk}} p(h_p = 1|x) x_q + \\ 
& \hspace{0.5cm} p(x)\sigma (\widetilde{net_p}) (1- \sigma(\widetilde{net_p})) x_q x_k  ], \end{aligned}
\end{equation*}
\begin{equation*}
\begin{aligned}
& \frac{\partial^2 RBM}{\partial w_{kq} \partial w_{pq}} \\
& = \frac{\partial}{w_{kq}} [p(h_p = 1|x) x_q)] -  \frac{\partial}{w_{kq}} [ \sum_x p(x) p(h_p = 1|x) x_q] \\
& \hspace{0cm} = - \sum_x [\frac{\partial p(x) }{\partial w_{kq}} p(h_p = 1|x) x_q + p(x) \frac{\partial }{\partial w_{kq}} [p(h_p=1|x)x_q ]  ], \\
& \frac{\partial^2 RBM}{\partial w_{kp} \partial w_{pq}} = - \sum_x [\frac{\partial p(x) }{\partial w_{kp}} p(h_p = 1|x) x_q + p(x)] \\
\end{aligned}
\end{equation*}
where $\sigma ()$ is the sigmoid function, $\widetilde{net_p}$ is $\sum_{q} w_{pq} x_q + c_p $, and $c_p$ is the hidden bias. Based on what we derive above we can calculate the gradient of approximated $|| [\nabla_{\theta_{DBN}} - log \: p(x)]_i ||^2$.

\end{document}